\documentclass[10pt,journal,compsoc]{IEEEtran}
\usepackage{epstopdf}
\usepackage{amssymb}
\usepackage{multirow}
\usepackage{graphics}
\usepackage{graphicx}
\usepackage{tabularx}
\usepackage{epstopdf}
\usepackage{graphicx, graphics, psfrag, epsfig, latexsym, amssymb, amsmath, pifont}
\begin{document}
\title{Deep Representation of Facial Geometric and Photometric Attributes for Automatic 3D Facial Expression Recognition}
\author{Huibin~Li,
Jian~Sun$^{*}$,
Dong~Wang, Zongben~Xu,
 and~Liming~Chen
\IEEEcompsocitemizethanks{\IEEEcompsocthanksitem
H. Li, J. Sun, D. Wang and Z. Xu are with the Institute for Information and System Sciences, School of Mathematics and Statistics, Xi'an Jiaotong University, Xi'an, 710049, China.
E-mail: {huibinli, jiansun,zbxu}@mail.xjtu.edu.cn, dwang.6257302@stu.xjtu.edu.cn.
\IEEEcompsocthanksitem
L. Chen is with the LIRIS UMR5205, Department of Mathematics and Informatics, Ecole Centrale de Lyon, Lyon, 69134, France.
E-mail: liming.chen@ec-lyon.fr.
}
\thanks{Manuscript received xxx, xxx; revised xxx, xxx.}
}
\IEEEtitleabstractindextext{%
\begin{abstract}
In this paper, we present a novel approach to automatic 3D Facial Expression Recognition (FER) based on deep representation of facial 3D geometric and 2D photometric attributes. A 3D face is firstly represented by its geometric and photometric attributes, including
the geometry map, normal maps, normalized curvature map and texture map. These maps are then fed into a pre-trained deep convolutional neural network to generate the deep representation. Then the facial expression prediction is simply
achieved by training linear SVMs over the deep representation for different maps and fusing these SVM scores.
The visualizations  show that the deep representation provides a complete and highly discriminative coding scheme for 3D faces.
Comprehensive experiments on the BU-3DFE database demonstrate that the proposed deep representation
can outperform the widely used hand-crafted descriptors (i.e., LBP, SIFT, HOG, Gabor)  and the state-of-art approaches under the same experimental protocols.
\end{abstract}
}
\maketitle
\IEEEdisplaynontitleabstractindextext
%
\IEEEpeerreviewmaketitle

\IEEEraisesectionheading{
\section{Introduction}
\label{sec:introduction}}
Facial expression is one of the most natural and preeminent way for human beings to express and communicate their emotions, opinions and intentions.
Its automatic analysis and recognition has a wide range of applications, such as smart Human Computer Interface (HCI), psychology studies, entertainment etc.~\cite{ZengPRH09},~\cite{PanticR00}. Thus, machine-based facial expression and emotion analysis, tracking, and recognition has been extensively investigated over the last two decades~\cite{PanticR00},~\cite{Sandbach2012},~\cite{Fang2011}.

Facial Expression Recognition(FER) methods generally can be classified from three perspectives:
i.e., the data modality, the expression granularity, and the temporal dynamics.
From the first perspective, they are classified into 1) 2D FER (which uses 2D gray or color face images),
2) 3D FER (which uses 3D range images, point clouds, or meshes of faces).
and 3) 2D + 3D multi-modal FER (which uses both 2D and 3D facial data).
From the second perspective, they are divided into
1) recognition of six basic facial human emotions, namely anger (AN), disgust (DI), fear (FE), happiness (HA), sadness (SA) and surprise (SU). 2) detection and recognition of facial Action Unit
(AU, e.g., brow raiser, lip tightener, and mouth stretch).
From the third perspective,
they are categorized into static (still images) and dynamic (image sequences)
FER. In this paper, we focus on the problem of recognizing the six basic facial
expressions using 2D and 3D static facial attribute maps.

In the past decades, the majority of methods dedicated to develop FER system based on 2D facial data~\cite{PanticR00}. Despite significant advances have achieved in 2D FER, it
still fails to solve the two major challenges: illumination changes and pose variations~\cite{PanticR00}.
Recently, with the rapid development of 3D imaging and scanning technologies,  it is more and more popular to capture face scans for 3D face recognition and 3D facial expression analysis.
3D facial data can provide complete and accurate facial geometry and topology structures, which are naturally robust to poses and lighting variations.
Thus, FER using 3D data (3D scans and 3D videos) has attracted wide attentions in recent years ~\cite{Sandbach2012},~\cite{Fang2011},~\cite{Fang2012738}.

 \begin{figure*}[t!]
\begin{center}
   \includegraphics[width=1.0\textwidth]{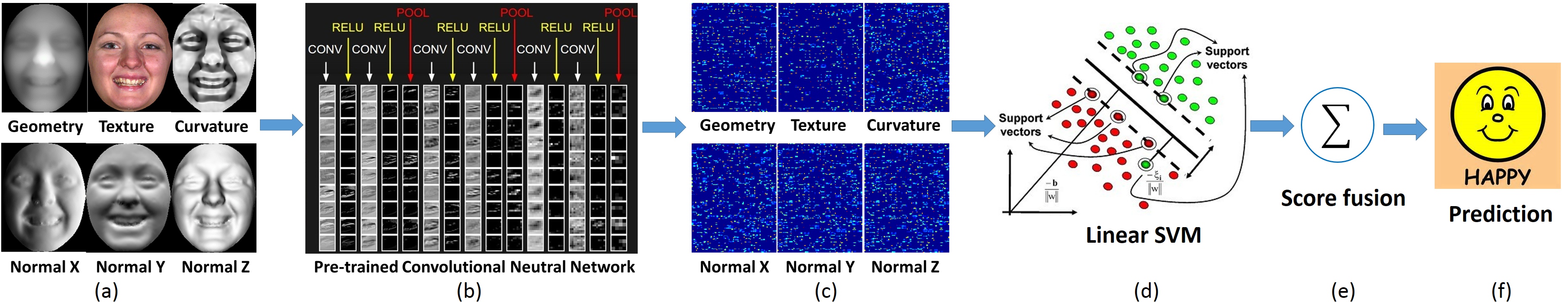}
\end{center}
\vspace*{-4 mm}
   \caption{Overview of the proposed approach. (a) Geometric and photometric attribute maps of a 3D face.
   (b) A pre-trained non-linear deep convolutional neutral network (the figure is copied from the Stanford CS class CS231n). (c) The deep representations of facial geometric and photometric information.
   (d-f) The linear SVM is trained and tested for score-level fusion and expression prediction.
    \label{fig:framework}}
\vspace*{-2 mm}
\vspace*{-3 mm}
\end{figure*}

\subsection{Related works}
3D FER methods can be roughly divided into two categories: \emph{model-based} and \emph{feature-based} approaches~\cite{Fang2011},~\cite{PanticR00}.
For \emph{model-based} approaches, dense rigid registration and non-rigid fitting techniques are utilized to get one-to-one point correspondence
among face scans, a generic expression deformable model is then generated to fit unknown faces, and their fitting parameters are finally used for
expression prediction. For example, a set of training faces of several persons depicting different facial expressions were used to establish an elastically deformable model in~\cite{TIFS2008}.
Then, asymmetric bilinear models of unknown faces were fitted to the eigenvectors of the deformable model and their expression and
identity control parameters were estimated. These parameters were used to build Maximum Likelihood classifier for final 3D FER.
Gong \emph{et al.}~\cite{gong2009automatic} suggested to learn a model to decompose the shape of an expressive face into a neutral-style basic facial shape component (BFSC) and an expression shape component (ESC).
The ESC is then used to design expression features.
In~\cite{ICPR2010Zhao}, Zhao \emph{et al.} proposed to build a Statistical Facial feAture Model (SFAM) for automatic landmarking, and extract shape and
texture features around landmarks for 3D FER.
The main drawback of model-based approaches lies in that they require
to establish dense correspondence among face scans, which is still a very challenging issue.
Meanwhile, time consuming procedures like dense 3D face registration and model fitting are indispensable.

\emph{Feature-based} approaches generally extract local expression features around facial landmarks based on
surface geometric attributes or differential quantities.
For example, 3D landmark distances~\cite{soyel2007facial},~\cite{soyel20083d},~\cite{tang20083dfg},~\cite{tang20083d},
3D curves~\cite{maalej2010local}, geometry and normal maps~\cite{ocegueda2011expressive}, conformal images ~\cite{zeng2013automatic},
surface normal~\cite{li20113d} and curvatures~\cite{li20113d},~\cite{wang20063d} are some popular 3D shape features.
\emph{Feature-based} approaches generally perform better than \emph{model-based} ones. However, their performances are largely depend on the accuracy of 3D facial landmarking, which is a challenging task~\cite{Fang2011}.

 \begin{figure*}[t]
\begin{center}
   \includegraphics[width=1.0\linewidth]{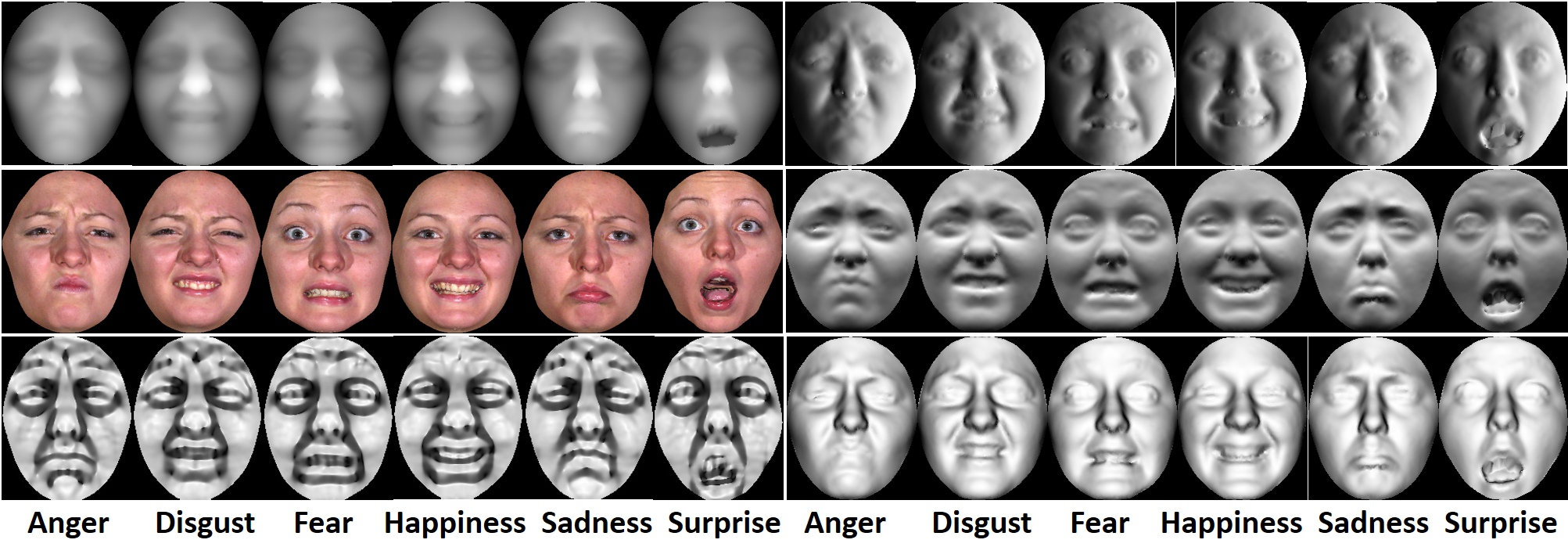}
\end{center}
\vspace*{-4 mm}
   \caption{The facial geometric and photometric attribute maps of 3D faces using six types of normalized 2D facial maps
   (i.e. geometry, texture, curvature, normal components $x$, $y$, and $z$) for six basic expressions of the subject F0001 in the BU-3DFE database.
    \label{fig:maps}}
\vspace*{-2 mm}
\end{figure*}

\subsection{Motivation and contributions}
Facial feature or face representation is the most important part for current state-of-the-art FER systems.
In fact, the visual descriptors like SIFT, HOG and LBP, that had huge impacts in the computer vision community,
have been successfully exploited for both 2D and 3D FER~\cite{2010ICPRBerretti},~\cite{lemaire},~\cite{li20123d}.
Recently, these hand-crafted descriptors have been substantially outperformed by automatically-learned \emph{deep features}
in extensive applications ranging from low-level restoration tasks~\cite{deepSuper},~\cite{deepdenoise}
to high-level object recognition tasks~\cite{Chatfield14},~\cite{NIPS2012_4824}.

Recently, Razavian \emph{et al.}~\cite{RazavianASC14} demonstrated a  simple yet effective recognition system using
the CNN features extracted from off-the-shelf pre-trained \emph{OverFeat} model~\cite{OverFeat}
and the linear SVM for recognition.
Surprisingly, this straight-forward recognition system consistently outperforms the finely tuned state-of-the-art systems on various databases for visual recognition tasks, such as image classification, scene recognition, fine-grained recognition, attribute detection and image retrieval.

Inspired by these results, the goal of this paper is to find further evidence of the success of CNN features for the task of 3D facial expression recognition.
Our aim is to design the \emph{deep representation} of 3D face scans for facial expression recognition.
We represent a 3D face scan by the deep features over its geometric and photometric attributes, and futher
evaluate and verify the advantages of deep facial representation on the benchmark 3D FER database.
In dosing so, we are interested in the following questions:
(i) Which pre-trained CNN net should be used for our 3D FER task?
(ii) Which layer of this net should be best for the deep representations of facial information?
(iii) What kinds of good properties does the deep representation have?
(iv) How about the performance of the deep representation compared to the state-of-art hand-crafted representations?

To answer these questions, in this paper, we thus employed several pre-trained CNN nets for the deep representations of multiple facial geometric and photometric attributes (Section~\ref{sec:attribute} and Section~\ref{sec:nets}).
We also visualized the fascinating deep representations to seek their secrets (Section~\ref{sec:vis1} and Section~\ref{sec:vis2}),
evaluated their performance using features in different net layers (Section~\ref{sec:layer}),
and compared them to the state-of-the-art descriptors and systems (Section~\ref{sec:score}).
Note that the \emph{deep feature learning} approach has been applied to 2D FER
~\cite{Shizhong},~\cite{Disentangling}, but we focus on 3D FER in this paper.

The major contributions of this paper are summarized as follows:
(i) We are the first to introduce the deep representation of 3D geometric (i.e. geometry, normal, and curvatures)
 and 2D photometric (i.e. texture) facial attributes to the automatic 3D FER.
(ii) By visualizing the deep representation of facial images, we find that the introduced deep representation
can provide us a complete and discriminative representation for facial attributes.
(iii) We conduct comprehensive experimental evaluations on the benchmark 3D FER database and report
consistent better results compared to the shallow representations (i.e. LBP, SIFT, HOG, Gobor)
as well as the state-of-the-art 3D FER systems.
Our results suggest that the proposed deep representation of 3D faces achieves state-of-the-art results in 3D FER task.

\vspace*{-2 mm}
\section{Overview of the proposed approach}
Figure~\ref{fig:framework} illustrates an overview of the proposed approach.
Given a raw textured 3D face scan, we first run the preprocessing pipeline in~\cite{mian2007efficient}, including nose tip detection, face cropping,
and pose normalization. The preprocessed textured 3D scan is then projected to 2D plane using mesh interpolation technique to generate
2D texture map $I_t$ and 2.5D geometry map $I_g$. The coordinates information of each geometry map are then used to estimate the surface normals and curvatures of this person, resulting in three normal component maps $I_n^x$, $I_n^y$, and $I_n^z$, and one normalized curvature (i.e. shape index) map $I_c$.
Finally, the geometric and photometric attributes of a textured 3D face scan $I$ can be  described by six types of 2D facial maps, i.e.
$I = \{I_g, I_t, I_c, I_n^x, I_n^y, I_n^z \}$, as shown in Fig.~\ref{fig:framework}~a.
Each of these maps is then fed into a pre-trained deep convolutional neural network  (CNN) (Fig.~\ref{fig:framework}~b) for
producing their corresponding deep representations: $\{f(I_g), f(I_t), f(I_c), f(I_n^x), f(I_n^y), f(I_n^z) \}$ (Fig.~\ref{fig:framework}~c), where $f$ is the non-linear mapping by CNN.
The linear SVM classifiers are trained and tested respectively for each type of those deep representations,
and their scores are fused with a simple \emph{sum rule}  for the final expression classification (Fig.~\ref{fig:framework}~d-f).

\section{Attribute Maps of a 3D face}
\label{sec:attribute}
To comprehensively describe the geometric and photometric attributes of a textured 3D face scan,
six types of 2D facial attribute maps, namely the geometry map $I_g$, texture map ($I_t$), three normal maps ($I_n^x$, $I_n^y$, and $I_n^z$),
as well as normalized curvature map ($I_c$) are employed. The geometry and texture maps are generated by performing the preprocessing and the interpolation-based mesh projection procedures on the raw face data. The normal and curvature maps are produced by estimating the normals and curvatures of 3D facial scans on the geometry map, which will be introduced as follows.
\subsection{Normal maps}
Given a normalized facial geometry map $ I_g$ represented by a $ m \times n \times 3$ matrix:
\begin{equation}
I_g = [p_{ij}(x,y,z)]_{m \times n} = [p_{ijk}]_{m \times n \times \{x,y,z\}},
\end{equation}
where $p_{ij}(x,y,z) = (p_{ijx},p_{ijy},p_{ijz})^T, (1\leq i \leq m, 1\leq j \leq n, i, j \in  \mathbb{Z})$  represents the 3D coordinates of point $p_{ij}.$
Let its unit normal vector matrix ($ m \times n \times 3$) be
\begin{equation}
I_n = [n(p_{ij}(x,y,z))]_{m \times n} = [n_{ijk}]_{m \times n \times \{x,y,z\}},
\end{equation}
where $n(p_{ij}(x,y,z)) = (n_{ijx},n_{ijy},n_{ijz})^{T}, (1\leq i \leq m, 1\leq j \leq n, i, j \in  \mathbb{Z}) $ denotes the unit normal vector of $p_{ij}. $
In this paper, we utilize the local plane fitting method \cite{NormalEstimation} to estimate $I_n$.
That is to say, for each point  $p_{ij} \in I_g$,  its normal vector $n(p_{ij})$ can be estimated as the normal vector of the following local fitted plane:
\begin{equation}
S_{ij}:  n_{ijx}q_{ijx}+n_{ijy}q_{ijy}+n_{ijz}q_{ijz} = d,
\end{equation}
where $(q_{ijx}, q_{ijy}, q_{ijz})^{T}$ represents any point within the local neighborhood of point $p_{ij}$ and $d= n_{ijx}p_{ijx}+n_{ijy}p_{ijy}+n_{ijz}p_{ijz}.$ In this work, a neighborhood of $5 \times 5$ window is used.
To simplify, each normal component in equation (2) can be represented by an $m \times n$ matrix:
\begin{equation}
I_n =
\begin{cases}
I_n^x = [n_{ij}^{x}]_{m \times n},\\
I_n^y = [n_{ij}^{y}]_{m \times n},\\
I_n^z = [n_{ij}^{z}]_{m \times n}.
\end{cases}
\label{eq:normal}
\end{equation}
where $\|(n_{ij}^{x},n_{ij}^{y},n_{ij}^{z})^T\|_{2} = 1.$
\subsection{Curvature map}
Similar to the local plane fitting method used for normal estimations,
we explored the local cubic fitting method~\cite{goldfeather2004novel} to estimate the principle curvatures.
This method assumes that the local geometry of a surface is approximated by a cubic surface patch.
For robustly solving the local fitting problem, both the 3D coordinates and the normal vectors of the
neighboring points of the point $p_{ij} \in I_g$ to be estimated are used.
That is, we are fitting the following equations:
\begin{equation}
\begin{cases}
z(x,y) =
\frac{a}{2}x^{2} + bxy + \frac{b}{2}y^{2} + dx^{3}+ ex^{2}y + fxy^{2} + gy^{3}\\
z_x = ax + by + 3dx^{2} + 2exy + fy^{2}\\
z_{y} = bx + cy + ex^{2}+2fxy + 3gy^{2}.
\end{cases}
\label{eq:normal}
\end{equation}
These equations can be solved by the least squares regression and the shape operator is computed as:
\[
W =
\begin{pmatrix}
  a & \quad b\\
  b & \quad c
\end{pmatrix}
.\]
 \begin{figure*}[t!]
\begin{center}
   \includegraphics[width=1.0\linewidth]{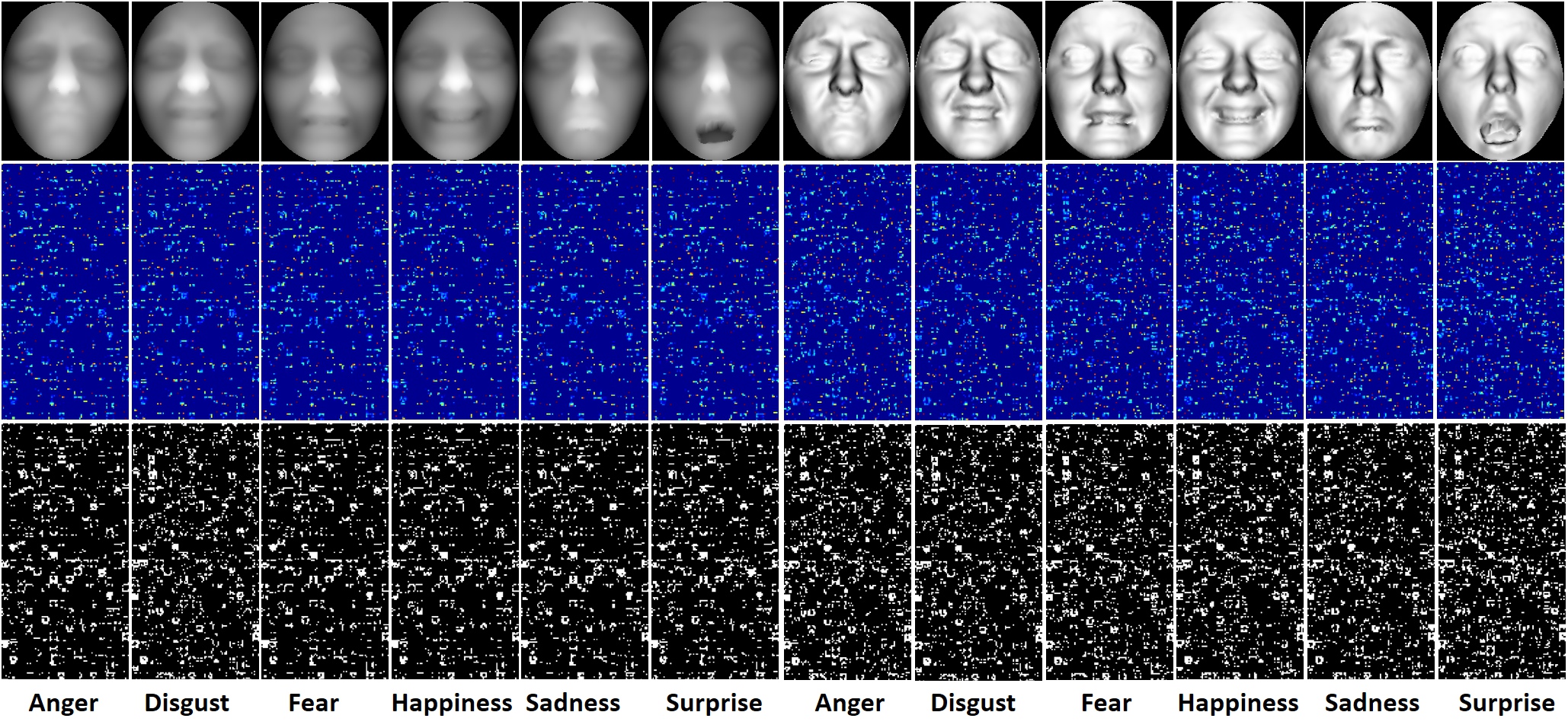}
\end{center}
\vspace*{-4 mm}
   \caption{Visualization of the deep representation of facial geometry maps $I_g$ (left, the first row) and normal component maps $I_n^z$ (right, the first row) of the same person with six basic expressions. The second row shows the color maps of the resized deep features (size from $6 \times 6 \times 512$ to $192 \times 96$). The values of these color maps are normalized to [0,1], where the blue indicates 0. And their corresponding binary code maps, achieved by resetting all the values bigger than 0 to 1, are illustrated in the third row.
    \label{fig:gpimages}}
\vspace*{-2 mm}
\end{figure*}
Then, the eignvalues of $W$ give the two principle curvatures $\kappa_1$ and $\kappa_2$ at point $p_{ij} \in I_g$.
The normalized curvatures (i.e. shape index value) at this point is defined by:
\begin{equation}
 \frac{1}{2} - \frac{1}{\pi}\arctan \bigg( \frac{\kappa_1 + \kappa_2}{\kappa_1 - \kappa_2} \bigg).
\end{equation}
Figure~\ref{fig:maps} shows the facial geometric and photometric attribute maps using six types of normalized 2D facial maps
for six expressions of a subject in the BU-3DFE database.

\section{Deep representation of a 3D face}
This section first gives a short introduction to the pre-trained CNN models used to generate the deep representation of a 3D face.
And then intuitively visualizes the coding patterns and the saliency maps of deep representation, from which
we discovered two powerful properties of deep representation for 3D face, i.e., \emph{completeness} and \emph{discriminative ability}.

\subsection{The architectures of pre-trained CNNs}
\label{sec:nets}

{\bf Caffe-Net} Two pre-trained models: \emph{caffe-ref} and \emph{caffe-alex}, implemented by the Caffe~\cite{jia2014caffe} software
are used. The architectures of the models are copied from the AlexNet~\cite{NIPS2012_4824}, which contain eight learned layers: five convolutional and three fully connected layers. The input of the AlexNet is a fixed-size $227 \times 227$ RGB image.

{\bf VGG-Net} Three pre-trained deep VGG-Net models~\cite{Chatfield14} with fast architecture (\emph{vgg-net-f}), medium architecture (\emph{vgg-net-m}),
and slow architecture (\emph{vgg-net-s}) and two pre-trained very deep VGG-Net models
: VGG Net-D (\emph{vgg-verydeep-16}) and VGG Net-E (\emph{vgg-verydeep-19})~\cite{Simonyan15} are used to generate deep representation in this paper.
The deep VGG-Net architectures consist of 5 convolutional layers and 3 fully-connected layers.
The differences between the fast, medium, and slow architectures involve
the number and receptive filed size of convolutional filters, convolution strides, spatial padding styles, and max-pooling down-sampling factors.
The fully-connected layers contain two 4,096 channel layers and one final 1,000 channel soft-max layer.
The very deep VGG-Net architectures contain 13 or 16 convolutional layers and 3 fully-connected layers, resulting in totally 16 (\emph{vgg-verydeep-16})
or 19 (\emph{vgg-verydeep-19}) weight layers. The sizes of convolutional filters are fixed to $3 \times 3$, and the numbers of filters in different layers are range from 64 to 512. The input of VGG-Net is a fixed-size $224 \times 224$ RGB image.
All these 7 pre-trained models are download from the research page of~\cite{Chatfield14}.



 \begin{figure*}[htbp!]
\begin{center}
   \includegraphics[width=1.0\linewidth]{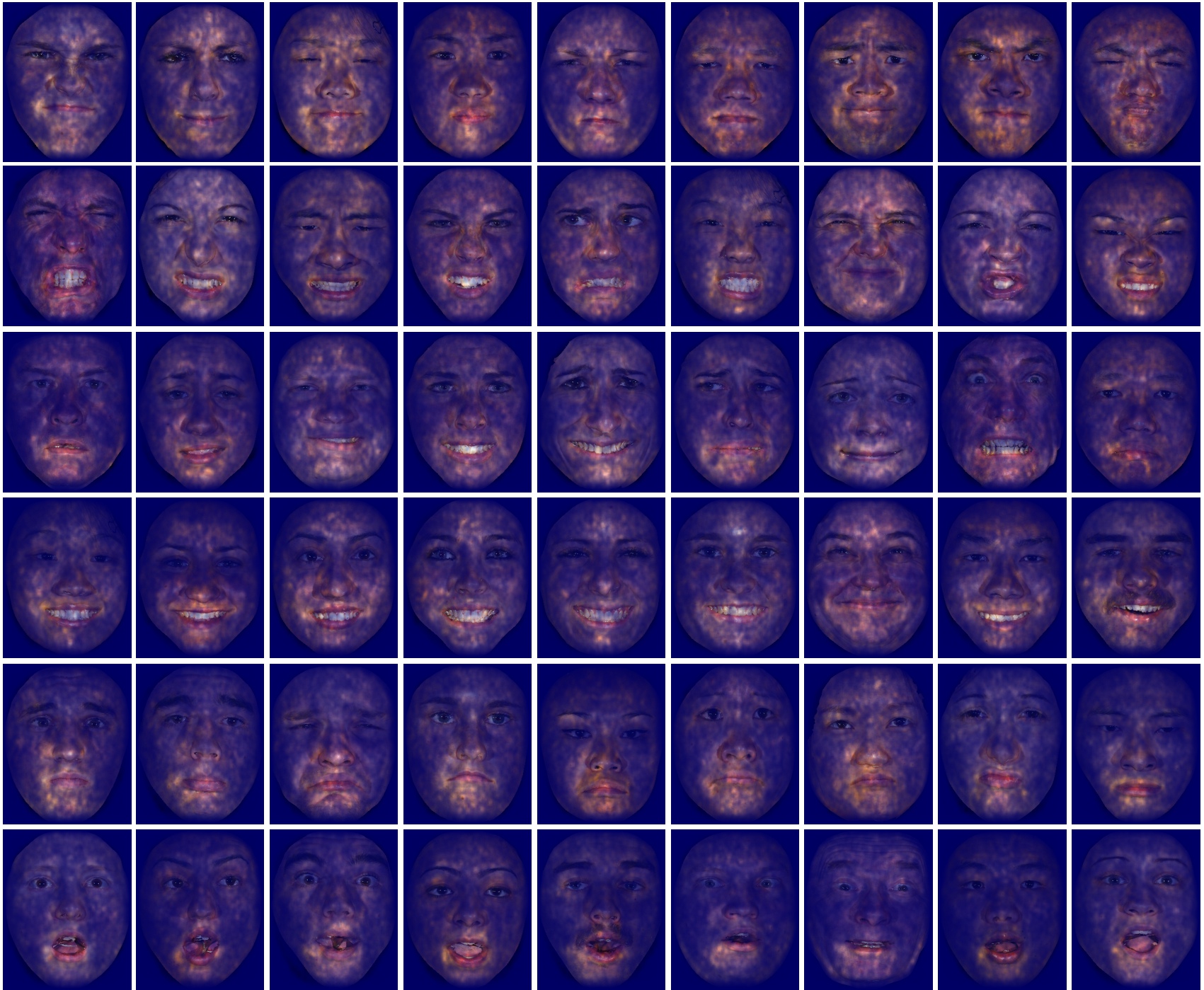}
\end{center}
\vspace*{-4 mm}
   \caption{Visualization of the saliency maps of deep representation with six different expressions: anger, disgust, fear, happiness, sadness and surprise (from top to down).  The less important pixels are shown in dark blue in these maps. The figures indicate that: 1) the mouses are the most salient parts for discriminating all these expressions. 2) The distributions of those salient parts for all expressions are consistent with the patterns of facial shape deformations, which may spread over the whole faces with different importance.
    \label{fig:Saliencymap}
    }
\vspace*{-4 mm}
\end{figure*}

\subsection{Deep representations and their visualization}
\label{sec:vis1}
A textured 3D face is represented by a set of normalized geometric and photometric attribute maps,
i.e., $I = \{I_g, I_t, I_c, I_n^x, I_n^y, I_n^z\}.$
Given a pre-trained CNN model $M_{cnn}$ introduced in Section~\ref{sec:nets},
the CNN hierarchically extracts multi-layer deep features $f(I_{i} \in I, M_{cnn})$ from each attribute map indexed by $I_i$ in $I$. In our approach, we use the deep features of the final pooling layer just after the final convolutional layer (the detail for layer selection
will be discussed in Section~\ref{sec:layer}),
which is a $m_{cnn} \times m_{cnn} \times d$ matrix, where $m_{cnn} \times m_{cnn}$ is the spatial resolution and $d$ is the number of feature channels
determined by the number of filters in the final convolutional layer.
Taking \emph{``vgg-net-s''} as an example, the deep feature matrices are with the size of $6 \times 6 \times 512$ dimensions for any facial map $I_i$.
We stack them into $18,432 \times 1$  dimensional feature vectors, and then perform $l_2$ normalization procedure.
These feature vectors are called the \emph{deep representation} of a textured 3D face scan,
and then used to learn the linear SVM classifiers for the final expression prediction.

These deep representations have intuitive meaning. Following the above example,
suppose the geometry map $I_g$ and normal component map $I_n^z$ of a textured 3D face are fed into the pre-trained CNN \emph{``vgg-net-s''}, respectively,
we resize their 3D deep representation matrices ($6 \times 6 \times 512$) into 2D deep representation matrices ($192 \times 96$) and visualize them in Figure~\ref{fig:gpimages}. It's easy to observe that the deep codes of faces are generally very sparse, and with very high dimensionality.
This implies that the deep CNN provides a high-dimensional feature space, and each facial attribute map can be represented by a sparse code, similar to the sparse representation~\cite{MairalBP14} or the widely used 2-dimensional bar code. This means that the CNN provides a complete (perhaps over-complete) representation of facial information, and we call this property as "completeness". This sparse coding scheme is essentially different to the traditional hand-crafted descriptors, like LBP, SIFT, etc.

\subsection{Saliency maps for deep representations}
\label{sec:vis2}
Deep representations also have very strong \emph{discriminative} ability, which can be discovered
by visualizing the saliency maps for different facial expressions.
 In this visualization approach, we aim to visualize the importance of each pixel for the final discrimination of facial expressions. For example, for the expression of ``happiness'', we visualize the saliency map for a 3D face by the importance of each image pixels contributing to the final discrimination of ``happiness''.

To compute the saliency map of a  3D face $I$ w.r.t. an expression indexed by $e$,
we construct a score function for assigning this face to expression $e$ by:
\begin{equation}
S(I | e, M_{cnn} ) = \sum_{{I_i \in {I}}} {{w_{I_i}^e}^T f(I_i, M_{cnn})},
\label{eq:score}
\end{equation}
where $M_{cnn}$ is a deep CNN model for extracting the deep features, $I_i$ is a geometric or photometric attribute map for 3D face $I$,  $f(I_i, M_{cnn})$ is the extracted deep feature vector for the input map. $w_{I_i}^e$ is the weights  of learned SVM classifier for expression $e$ using the training attribute map of $I_i$. Obviously, the higher value of ${{w_{I_i}^e}^T f(I_i, M_{cnn})}$ implies higher confidence in labeling this 3D face as expression $e$, the final score  fuses these confidences from different attribute maps.
We next compute the gradient of score function in Eqn.~(\ref{eq:score}) w.r.t. the input pixels:
\begin{equation}
G(x |I, e, M_{cnn}) = \sum_{I_i \in I _{\Lambda}} {{w_{I_i}^e}^T  \frac{\partial f(I_i, M_{cnn})}{\partial I_i(x)}}
\end{equation}
where $x$ is a pixel on the 3D face.  $\frac{\partial f(I_i, M_{cnn})}{\partial I_i(x)}$ is the gradient of deep feature w.r.t. the attribute map $I_i$ at pixel $x$, which can be computed by back-propagation. Its absolute value $|G(x |I, e, M_{cnn})|$ measures the importance of pixel $x$ in labeling $I$ as expression $e$.  We call this term computed over all 3D face pixels as \emph{saliency map.}

Figure~\ref{fig:Saliencymap} visualizes some examples of saliency maps for different expressions. The saliency map is re-scaled to [0, 1]. We visualize it by fusing the face texture map with a dark blue background using the saliency map as weights. The less important pixels are shown in dark blue in these maps. We observe some interesting phenomena from these maps. First, the mouth are the most salient parts for discriminating all these expressions of interest. Second, the distributions of those salient parts for all expressions are consistent with the patterns of facial shape deformations, which may spread over the whole faces
with different importance. Thus, the deep CNN provides a discriminative representation for 3D face by distinguishing facial expressions using the discriminative facial parts.

\section{Experimental results}
\subsection{The BU-3DFE dataset}
We evaluate our deep representation-based 3D FER approach on the BU-3DFE \cite{BU3FE} dataset.
It consists of 2,500 textured 3D face scans of 100 subjects (including 56 females and 44 males) with different ages and races.
Each subject has six basic expressions: anger (AN), disgust (DI), fear (FE), happiness (HA), sadness (SA), and surprise (SU)
with 4 intensity levels plus a neutral (NE) expression, a total of 25 samples.
\subsection{Experimental settings}
For fair comparison, two common experimental protocols are used.
In both protocols, 720 3D face scans of 60 subjects displaying the six basic expressions with intensity of level 3 and level 4 are utilized.
For Protocol I, it is required to perform 100 times 10-fold cross-validation, and  60 subjects are randomly selected
for training and testing at each time~\cite{2010ICPRBerretti}.
For Protocol II, it is required to run 100 times 10-fold cross-validation, but the 60 subjects are randomly selected once and then fixed at all times for training and testing \cite{gong2009automatic}.
That is, 60 subjects are randomly divided into 10 subsets, 9 subsets (54 subjects with 648 samples) are used for training, while the remaining 1 subset (6 subjects with 72 samples) are used for testing. The experiments are repeated 10 times and the average recognition score from the 10 splits is used as the performance of a round 10-fold cross-validation.

As introduced in Section~\ref{sec:nets}, we evaluate two types of nets: Caffe-Net, Vgg-Net and their variants, a total of 7 nets for performance evaluation.
Four hand-crafted image descriptors: MS-LBP, dense-SIFT, HOG, and Gabor are used for comparison.
Please refer to~\cite{huibinbtas},~\cite{vedaldi08vlfeat},~\cite{lemaire},~\cite{huibinbtas}, respectively for the source codes or parameters of these descriptors.
The features are $L_2$ normalized and then fed into the linear SVM~\footnote{http://www.csie.ntu.edu.tw/~cjlin/liblinear/} with  hinge loss and $L_2$ regularization. The parameter $C$ is set to 1.
Score-level fusion with simple sum rule is used for all the fusion procedures.


\subsection{Layer selection for deep representation}
\label{sec:layer}
To verify which layer of a pre-trained deep CNN should be best to extract the deep representation of a 3D face,
we generate layer-dependent deep representations for different facial attribute maps and compare their accuracies for 3D FER.
In particular, three representative pre-trained nets, namely \emph{caff-alex},  \emph{vgg-net-m}, and  \emph{vgg-verydeep-16}
are used, and their results are reported in Table~\ref{tab:caffenet}, Table~\ref{tab:vggnetm}, and Table ~\ref{tab:vggnetdeep16}, respectively.
All these results are achieved by running 10 times of 10-fold cross-validation (i.e., 100 times training and testing sessions) by Protocol I.
In practice, for computational efficiency, we choose the deep features of a convolutional layer following a max-pooling layer for all nets.

\begin{table}[htbp!]
\begin{center}
\resizebox{0.48 \textwidth}{!}{
\begin{tabular}{|l|c|c|c|c|c|c|c|}
\hline
\%& conv1 & conv2 &conv3 &conv4 &conv5 &full6 &full7\\
\hline
dim&69,984&43,264 &64,896&64,896 &9,216 &4,096 &4,096\\
\hline
$I_g$&77.10&{\bf 79.81}&79.60&79.61&76.28&72.25&66.24\\
$I_n^x$&78.65&{\bf 80.43}&79.43&79.32&78.25&76.06&72.11\\
$I_n^y$&79.40&{\bf 81.99}&81.82&81.76&80.76&78.08&74.76\\
$I_n^z$&79.89&{\bf 81.46}&80.75&80.38&78.63&76.47&73.79\\
$I_c$&77.29&80.21&{\bf 80.69}&80.00&77.81&74.87&70.38\\
$I_t$&75.64&{\bf 81.18}&81.11&80.68&79.51&77.29&73.17\\
\hline
$All$&81.21&{83.44}&{\bf 83.58}&83.38&82.93&81.64&80.49\\
\hline
\end{tabular}}
\end{center}
\caption{Dimension and performance comparison of  deep representations extracted from different attribute maps by different \emph{caffe-alex} net layers,
i.e., 5 convolutional layers (conv 1-5) and 2 fully connected layers (full 6-7).
}
\label{tab:caffenet}
\end{table}

\begin{table}[htbp!]
\begin{center}
\resizebox{0.48 \textwidth}{!}{
\begin{tabular}{|l|c|c|c|c|c|c|c|}
\hline
\%& conv1 & conv2 &conv3 &conv4 &conv5 &full6 &full7\\
\hline
dim&279,936&43,264&6,4896&86,528&18,432 &4,096 &4,096\\
\hline
$I_g$&  76.40&{\bf 80.31}&79.83&79.68&79.22&76.78&72.13\\
$I_n^x$&77.26 &{\bf 81.06}&80.31&80.21&78.93&77.00&72.79\\
$I_n^y$& 78.13&{\bf 81.49}&81.47&81.39&79.71&77.39&75.01\\
$I_n^z$& 78.88&{\bf 81.46}&80.97&80.57&79.71&77.43&74.69\\
$I_c$& 75.83&{\bf 80.36}& 79.96&79.10&76.99&72.56&66.71\\
$I_t$& 73.72&81.01&{\bf 81.22}&80.79&81.03&76.82&69.58\\
\hline
$All$&80.13&83.14&{83.07}&83.24&{\bf 83.50}&82.44&80.46\\
\hline
\end{tabular}}
\end{center}
\caption{Dimension and performance comparison of  deep representations extracted from different attribute maps by different
\emph{vgg-net-m} net layers,
i.e., 5 convolutional layers (conv 1-5) and 2 fully connected layers (full 6-7).
}
\label{tab:vggnetm}
\end{table}

\begin{table}[htbp!]
\begin{center}
\resizebox{0.48 \textwidth}{!}{
\begin{tabular}{|l|c|c|c|c|c|c|c|}
\hline
\%& conv1-2&conv2-2&conv3-3&conv4-3&conv5-3& full6 & full7\\
\hline
dim&802,816&401,408&200,704&100,352&25,088&4,096 &4,096\\
\hline
$I_g$&65.69&75.65&79.83&{\bf 81.18}&79.90&77.68&73.11\\
$I_x$&64.54&72.22&79.01&{\bf 80.78}&78.38&74.19&69.46\\
$I_y$&74.15&76.82&79.93&{\bf 82.11}&80.25&76.58&73.06\\
$I_z$&74.03&77.10&80.51&{\bf 81.51}&78.99&75.54&71.97\\
$I_c$&69.79&72.78&79.68&{\bf 80.93}&74.17&68.17&62.67\\
$I_t$&63.79&72.64&79.58&{\bf 80.99}&76.40&70.83&64.42\\
\hline
$All$&74.51&78.71&82.28&{\bf 83.97}&83.64&82.89&81.40\\
\hline
\end{tabular}}
\end{center}
\caption{Dimension and performance comparison of  deep representations extracted from different attribute maps by different \emph{vgg-verydeep-16} net layers,
i.e., 5 convolutional layers (conv 1-5) and 2 fully connected layers (full 6-7). In the table, convi-j means j-th sub-convolutional layer of i-th convolutional layer.
}
\vspace{-5mm}
\label{tab:vggnetdeep16}
\end{table}

\begin{table*}[htbp!]
\begin{center}
\begin{tabular}{|l|c|c|c|c|c|c|c|c|c|c|}
\hline
Method & $I_g$ & $I_n^x$ & $I_n^y$ & $I_n^z$ & $I_c$ & $I_t$  & $I_n$ & $I_g+I_n$ & $I_g+I_n+I_c$& $I_g+I_n+I_c+I_t$\\
\hline\hline
MS-LBP& 75.80&76.19&76.60&75.04&74.32&68.10&78.48&79.45&79.42&79.60\\
dense-SIFT& 78.63 & 79.16& 79.93& 79.72& 78.24& 73.47& 81.21& 81.22& 81.19&81.48\\
HOG&{\bf 79.67}&{\bf 79.82}&79.15&{\bf 80.14}&77.28&75.16&{\bf 81.43}&{\bf 82.01}&{\bf 81.72}&82.01 \\
Gabor&75.86&77.46&{\bf 80.49}&79.44&{\bf 79.12}&{\bf 76.78}&80.86&81.00&81.67& {\bf 82.71} \\
\hline
caffe-ref& 77.43  &79.16& 79.20 &78.30 & 77.82 & 78.19 &  80.40& 80.62 & 81.70& 82.62\\
caffe-alex & 76.51  &77.73& {\bf 80.16} &78.55 & 78.16 & { 79.32} &  80.13& 80.71 & 81.66& 82.63\\
vgg-net-f&77.62&77.68&79.46&78.77& 76.13 &{\bf 80.15}&80.40 &81.13&81.93&82.70\\
vgg-net-m&79.32&78.63&79.61&79.54&76.95&79.79&80.97&81.58 & 82.37 & 83.20 \\
vgg-net-s& {\bf 80.68}&{\bf 79.27}& 80.07& {\bf 80.34}&{\bf 79.04}&79.86&80.73& 81.37&82.24&83.07\\
vgg-verydeep-16 & 79.96  &77.29& 79.87 &79.21& 74.61 & 75.59  & 81.19 &{\bf 82.23}&82.69&83.31\\
vgg-verydeep-19 & 78.12 & 78.12 &78.89 &79.58& 74.31 & 75.62  & {\bf 81.68}&82.22&{\bf 82.81}& {\bf 83.48}\\
\hline
\end{tabular}
\end{center}
\caption{Average expression recognition accuracies of hand-crafted descriptors (i.e. MS-LBP, dense-SIFT, HOG, Gabor)
and deep representation of pre-trained nets. Results for each single attribute map and the score-level fusion are reported. Results are achieved by performing 100 times 10-fold cross-validation, at each time 60 subjects are randomly selected for training and testing (Protocol I).
}\label{tab:Protocol1}
\end{table*}

\begin{table*}[htbp!]
\begin{center}
\begin{tabular}{|l|c|c|c|c|c|c|c|c|c|c|}
\hline
Method & $I_g$ & $I_n^x$ & $I_n^y$ & $I_n^z$ & $I_c$ & $I_t$  & $I_n$ & $I_g+I_n$ & $I_g+I_n+I_c$& $I_g+I_n+I_c+I_t$\\
\hline\hline
MS-LBP&76.47&76.77&77.87&76.41&77.70&71.65&79.46&81.29&81.58&81.74\\
dense-SIFT&80.29&79.97&{\bf 82.35}&80.95&80.28&75.56&82.62&82.41&82.60&83.16\\
HOG&{\bf 81.89}&{\bf 82.09}&80.58&{\bf 81.81}&77.95&78.11&{\bf 82.72}&{\bf 83.62}&{\bf 83.28}&83.74 \\
Gabor&77.95&78.80&81.97&81.10&{\bf 81.65}&{\bf 80.36}&81.89&82.09&83.25& {\bf 84.72} \\
\hline
caffe-ref& 79.34  &80.05& 80.70 &78.69 & 80.38 & 78.97 &  80.79& 81.30 & 82.54& 83.46\\
caffe-alex & 77.53  &78.87&  81.50 &78.71 & {\bf 80.83} & { 81.40} &  81.05& 81.29 & 83.02& 83.74\\
vgg-net-f&79.67&79.25&81.66&79.82& 79.05 &{\bf 82.21}&81.58 &82.04&83.42&84.28\\
vgg-net-m&80.38&80.37&81.68&81.23&79.23&82.14&82.45&82.58 & 83.46 & 84.22 \\
vgg-net-s&  80.63&{\bf 81.70}& 80.94&  81.90 &80.16&81.44&81.92& 82.19&83.16&83.91\\
vgg-verydeep-16 & {\bf 81.72}  &78.55& {\bf 83.06} &81.25& 76.95 & 78.46  & 82.60 & 82.86&83.48&83.78\\
vgg-verydeep-19 & 81.04 & 80.25 &81.10 &{\bf 82.35}& 77.33 & 79.74  & {\bf 83.58}&{\bf 83.17}&{\bf 83.96}& {\bf 84.87}\\
\hline
\end{tabular}
\end{center}
\caption{Average expression recognition accuracies of hand-crafted descriptors (i.e. MS-LBP, dense-SIFT, HOG, Gabor)
and deep representation of pre-trained nets. Results for each single attribute map and the score-level fusion are reported. Results are achieved by performing 100 times 10-fold cross-validation, at all time fixed 60 subjects are randomly selected for training and testing (Protocol II).
}\label{tab:Protocol2}
\end{table*}

From Table~\ref{tab:caffenet}, Table~\ref{tab:vggnetm}, and Table~\ref{tab:vggnetdeep16},
we observe that the dimensions of the deep representations extracted from the convolutional layers
are much higher than the ones from the fully connected layers.
For example, the dimensions of the first convolutional layer (i.e., conv1) and the last fully connected layer (i.e., full7) are
69,984 vs. 4096, 279,936 vs. 4096, 802,816 vs. 4096 for \emph{caff-alex},  \emph{vgg-net-m}, and  \emph{vgg-verydeep-16}, respectively.
However, the convolutional layers (except conv1) generally perform much better than the fully connected layers for all attribute maps.
More precisely, the second convolutional layers (i.e., conv2) of  \emph{caff-alex} and \emph{vgg-net-m},
and the forth convolutional layer (i.e., conv4-3) of  \emph{vgg-verydeep-16} achieve the best results
for nearly all the attribute maps.
Note that there are two convolutional layers between conv2 and conv5 for \emph{caff-alex} and \emph{vgg-net-m}.
Similarly, there are also two convolutional (or sub-convolutional) layers between  conv4-3 and  conv5-3 for \emph{vgg-verydeep-16}.
These results suggest us to make a trade-off between the accuracy and the efficiency.

Notice that the last convolutional layer (i.e., conv5) has the same level of deep representation dimensions (around 2$\times$4,096, 4$\times$4,096, and 6$\times$4,096) compare with the fully connected layers (i.e., full6 or full7). Meanwhile, the performance of the conv5 are much better than the ones of full6 and full7. Moreover, when considering the score-level fusion of all the attribute maps, the accuracies achieved by conv5 are also very close to or even better than (for \emph{vgg-net-m}) the ones obtained by other convolutional layers.
Considering the balance between the accuracy and the efficiency, we suggest to choose the last convolutional layer (i.e. conv5) to extract the deep representation of 3D face scans for the three representative nets   \emph{caff-alex}, \emph{vgg-net-m}, and \emph{vgg-verydeep-16}.
In consideration of the strong structure similarity among these representative nets and the others, we also suggest to use the last convolutional layer (i.e. conv5) to extract the deep representation of 3D face scans for other four pre-trained nets (i.e.,  \emph{caff-ref}, \emph{vgg-net-f}, \emph{vgg-net-s}, and \emph{vgg-verydeep-19}) used in this paper.

\subsection{Performance evaluation and comparison}
\label{sec:score}
We now compare the performance of deep features to the popular hand-crafted features in vision community for 3D FER.
Table 1 and Table 2 show the average recognition accuracies of four hand-crafted descriptors and seven pre-trained nets in Protocol I and Protocol II, respectively. For deep representation, we present the average accuracies for both the separate attribute maps
and their score-level fusions. Based on these results shown in Tables~\ref{tab:Protocol1} and ~\ref{tab:Protocol2}, we have the following four conclusions:

(1) For hand-crafted descriptors, Gabor and HOG perform better than dense-SIFT and MS-LBP.
When considering the contributions of all attribute maps, Gabor achieves the highest accuracies
of 82.71\% and 84.7\% for Protocol I and Protocol II, respectively.

(2) For pre-trained nets, \emph{vgg-net-s}, \emph{vgg-net-m}, and \emph{vgg-net-f} are comparable to each other, which are slightly better than \emph{caffe-ref} and \emph{caffe-alex}, and \emph{vgg-verydeep-19} obtains the best results (83.48\% and 84.87\% for Protocol I and Protocol II, respectively) when considering the contributions of all attribute maps. This result is consistent with the main findings in~\cite{Simonyan15}, i.e., better results are
benefit from deeper net structure.

(3) For different attribute maps, normal maps perform slightly better than others and their fusion can generally improve the accuracy.
Moreover, the fusion of all 3D geometric attribute maps (i.e., $I_g + I_n + I_c$) generally performs better than the 2D photometric attribute map (i.e., $I_t$),
and their fusion (i.e., $I_g + I_n + I_c + I_t$) achieves the best performance for all nets. These results indicate that different facial attribute maps contain large complementary information for 3D FER.

(4) The results of the pre-trained nets are generally better than the results of hand-crafted descriptors, the superiorities are more obvious under the more fair comparison setting, i.e. Protocol I. Moreover, the results of Protocol II are better than the results of Protocol I, which indicates that the setting of Protocol II
contain considerable subject bias and thus are not as faithful as Protocol I.

Table~\ref{tab:confusionmatrix} compares the average confusion matrix of Gabor and the proposed deep representation
based on \emph{vgg-verydeep-19} in Protocol I. It can be seen that the proposed deep representation outperforms Gabor
for all expressions except sadness and surprise with slight differences.
In particular, deep representation have more powerful discriminative ability to distinguish fear expression.

\begin{table}
\begin{center}
\begin{tabular}{|l|cccccc|}
\hline
\% & AN & DI & FE& HA& SA & SU\\
\hline
Gabor&  &  & & &  & \\
AN & { 80.01}& 4.76 & 2.62 & 0.48 & 11.94 & 0.19\\
DI & 4.96 & { 81.37} & 3.66 & 2.85 & 2.72 & 4.44\\
FE & 4.51& 5.45 & { 62.84} & 14.15 & 5.64 & 7.41\\
HA & 0& 0.43 &3.93 & 95.17 & 0 & 0.48\\
SA & 14.56& 0.57 & 3.35 & 0.96 & 80.54 & 0\\
SU & 0.12& 0.96 & 1.38 & 0.95 & 0.24 & { 96.34}\\
\hline
deep &  &  & & &  & \\
AN & {\bf 82.10}& 4.58 & 2.57 & 0.48 & 10.27 & 0\\
DI & 3.64& {\bf 82.21} & 4.67 & 2.60 & 2.64 & 4.24\\
FE & 3.27& 7.68 & {\bf 68.19} & 11.98 & 4.72 & 4.17\\
HA & 0& 0 & 4.14 & {\bf 95.27} & 0 & 0.58\\
SA & 18.47& 1.14 & 2.87 & 0.34 & 77.17 & 0\\
SU & 0& 0.67 & 2.45 & 0.90 & 0 & 95.97\\
\hline
\end{tabular}
\end{center}
\caption{Comparison of the average confusion matrix between Gabor descriptor and deep representation based on \emph{vgg-verydeep-19} in Protocol I.
}
\label{tab:confusionmatrix}
\end{table}

Table \ref{tab:Comparison} reports the comparison results of the proposed method and the state-of-the-art 3D FER methods in the literatures of
~\cite{wang20063d},~\cite{soyel2007facial},~\cite{tang20083d}, ~\cite{gong2009automatic},~\cite{2010ICPRBerretti},~\cite{li20113d},~\cite{li20123d},
~\cite{lemaire},~\cite{zeng2013automatic},~\cite{zhen3dfer},~\cite{Yang3dfer}.
The early works ~\cite{wang20063d},~\cite{soyel2007facial},~\cite{tang20083d},~\cite{2010ICPRBerretti},~\cite{li20113d}
 except~\cite{gong2009automatic} require a large set of manual facial landmarks, marking them not automatic and not practical.
 Recent studies use holistic registration \cite{li20123d},~\cite{lemaire},~\cite{zhen3dfer},~\cite{Yang3dfer}
 or several automatically detected landmarks ~\cite{zeng2013automatic}, making their methods to be fully automatic.
 Benefit from the two powerful properties of deep representation for 3D face, i.e., completeness and discriminative ability,
 our method achieves an average recognition rate of 83.48\% and 84.87\% on the BU-3DFE dataset using Protocol I and Protocol II,
 which outperforms the state-of-the-art ones. Note that most of the state-of-the-art methods use SVM with non-linear RBF kernel~\cite{li20113d},~\cite{zeng2013automatic},~\cite{zhen3dfer},~\cite{Yang3dfer} or multiple kernel learning~\cite{li20123d}, but we only use linear SVM based on a single type of deep feature.

\begin{table}
\begin{center}
\begin{tabular}{|l|c|cc|}
\hline
\multicolumn{1}{|l}{\multirow{2}{*}{Method}}&\multicolumn{1}{|c}{\multirow{2}{*}{Automatic}}&\multicolumn{2}{|c|}{\multirow{1}{*}{Average acc. (\%) }}\\
\cline{3-4}
&  & Protocol I &Protocol II \\
\hline
2006 Wang et al. \cite{wang20063d}& No&- &61.79\\
2007 Soyel et al. \cite{soyel2007facial}&No &-&67.52\\
2008 Tang et al. \cite{tang20083d}& No & -&74.51\\
2009 Gong et al. \cite{gong2009automatic}&Yes  &-&76.22\\
2010 Berretti \cite{2010ICPRBerretti}& No& 77.54&- \\
2011 Li et al. \cite{li20113d}& No &- &82.01  \\
2012 Li et al. \cite{li20123d}& Yes & 80.14  & -  \\
2013 Lemaire et al.  \cite{lemaire} &Yes  &-&76.61\\
2013 Zeng et al.  \cite{zeng2013automatic} &Yes & 70.93 & -\\
2015 Zhen et al.  \cite{zhen3dfer} &Yes &  83.20 & 84.05\\
2015 Yange et al.  \cite{Yang3dfer} &Yes & 82.73 & 84.80\\
Ours (vgg-verydeep-19) &Yes & {\bf 83.48} & {\bf 84.87}\\
\hline
\end{tabular}
\end{center}
\caption{Comparison of the average recognition accuracy of the proposed method (deep representation of attribute maps based on \emph{vgg-verydeep-19}
and linear SVM) and the state-of-the-art ones.}
\label{tab:Comparison}
\end{table}

\section{Conclusion and future work}
In this paper, we propose a novel method for 3D FER.
Geometric and photometric attribute maps are employed to describe textured 3D face scans.
Deep representation of these attribute maps are then generated using deep convolutional
neural networks pre-trained on the large scale image classification task.
The visualization, net selection, layer selection, and performance of the proposed deep representation are studied.
We find that the deep representation of 3D face have two powerful properties: completeness and strong discriminative ability.
Experimental results on BU-3DFE data show that this kind of ``off-the-shelf'' deep representation can outperform the well known
hand-crafted descriptors (i.e., LBP, SIFT, HOG, Gobor) and the state-of-the-art methods for 3D FER.

In the future, we will study the fine-tuning technique to transfer the expression feature and label information to the pre-trained deep networks.
Moreover, we will also try to build a new deep CNN model for 3D FER by end-to-end training of the feature extraction in separate attribute maps, feature dimensionality reduction and fusion in a single framework. We are also interested in designing a 3D face deep representation based on facial landmarks.

%

\section*{Acknowledgment}
This work was supported in part by the National Natural Science Foundation of China (NSFC) under Grant
No.11401464 and No.61472313, Chinese Postdoctoral Science Foundation under Grant No. 2014M560785, and Agence Nationale de Recherche (ANR) through the Jemime project under grant No. ANR-13-CORD-004-02.


%
%

\ifCLASSOPTIONcaptionsoff
  \newpage
\fi


\bibliographystyle{ieee}
\bibliography{reftac}
\end{document}